\documentclass[11pt]{article}
\usepackage{geometry}                
\usepackage{graphicx}
\usepackage{amssymb}
\usepackage{epstopdf}
\usepackage{subfigure}
\title{Searching for Globally Optimal Functional Forms for Inter-Atomic Potentials Using Parallel Tempering and Genetic Programming.}
\author{A. Slepoy, A. P. Thompson, M. D. Peters}
\begin{document}
\maketitle
\abstract{ \it
We develop a Genetic Programming-based \cite{KozaJR:Genprh} methodology that enables discovery of novel functional forms for classical inter-atomic fields, used in molecular dynamics simulations. Unlike previous efforts in the field, that fit only the parameters to the fixed functional forms, we instead use a novel algorithm to search the space of many possible functional forms. While a follow-on practical procedure will use experimental and {\it ab inito} data to find an optimal functional form for a force-field, we first validate the approach using a manufactured solution. This validation has the advantage of a well-defined metric of success. We manufactured a training set of atomic coordinate data with an associated set of global energies using the well-known Lennard-Jones inter-atomic potential. We performed an automatic functional form fitting procedure starting with a population of random functions, using a genetic programming functional formulation, and a parallel tempering Metropolis-based optimization algorithm. Our massively-parallel method independently discovered the Lennard-Jones function after searching for several hours on $100$ processors and covering a miniscule portion of the configuration space. We find that the method is suitable for unsupervised discovery of functional forms for inter-atomic potentials/force-fields. We also find that our Parallel Tempering Metropolis-based approach significantly improves the optimization convergence time, and takes good advantage of the parallel cluster architecture.}
 
\section{ Introduction }
Classical Molecular Dynamics [MD] and other molecular mechanics simulations have become important computational tools in the nano-scale design of novel materials, and have brought insight into structure and function of bio-molecules. All such simulations require accurate and computationally efficient forms for the inter-atomic potential/force function. Force-fields that serve that role are typically physics-intuition-based functions of inter-atomic distances and bonding structure. Classical force-field functions represent extensive work in invention and validation, using physical insight combined with experimental and other sources of information. Computationally, construction of a force-field can be viewed as a fitting process, where successful functional forms of the force-field best "fit" the available data. Given a clear metric of "fitness", fitting a classical force-field is a well-defined optimization procedure. It requires a definition of a mutable representation for the function and parameters, a quantifiable "fitness" criterion, and a set of ergodic evolution operators. 

Until now, all the functional forms for such functions were generated by physical intuition, and only their parameters (i.e. multiplicative constants, additive constants, and exponents) were optimized. Each such process, done manually, represents many man-years of highly qualified labor, and often meets with failure. This important and laborious task clearly calls for automation. 

In the case of parameter-only fitting, a useful representation for the problem is an ordered set of real numbers, with each number acting as a parameter for a fixed functional form. The evolution operators, in this case, are generalized linear transformations that treat the ordered set of numbers as a vector in real space and search for the vector's optimal size and direction. A number of research efforts have attacked the partial problem of automatically fitting the numerical parameters in the fixed functional forms, successfully obtaining better fits to the objective function based on a training set.

As we become increasingly interested in complex multi-species systems, and look more critically to the quantitative prediction of their material properties, stringent requirements lead to the need for more complicated functional forms. Optimization of the functional form itself requires a substantial change in the methodology. Such more complicated functions are embedded in a very large combinatorial space. Researchers find it difficult, if not impossible, to develop an intuition for the relationships between the various functions and the ever-expanding training set. An automatic functional form fitting method is clearly called for. 

We use the Genetic Programming formalism as a starting point for such a fitting algorithm. Genetic Programming [GP]  \cite{KozaJR:Genprh} uses a library of elementary operators to build a population of hierarchical computer programs, represented as operator trees. We find that this description of the functional form provides us with maximum generality of the functional expression. We choose a minimal set of algebraic operators: $\{+,-,*,/, \hat {},| | \}$. If only the algebraic operators are used, the resulting tree becomes an algebraic expression with variable or constant number input at the leaves. The population of such trees is evolved using a Genetic Algorithm [GA] to improve the fitness of the tree population according to a user-defined fitness criterion. A small set of mutation operators provides a way to evolve the trees in the functional space. We describe the model in more detail in section \ref{subsec: GP}.

In the fitting problem, the training set, which underlies the fitness function, requires careful definition, robust design, and much labor. The objective function is typically based on a set of sets of atomic positions with an associated set of observables, obtained either experimentally or via { \it ab initio} simulations. For validation purposes, we construct a manufactured training set using an existing classical force-field. Such a manufactured solution provides the advantage of a clearly defined objective in terms of a known answer. We leave the somewhat philosophical discussion of the difference between the manufactured and a real system until section \ref{sec: Discussion}. We choose the Lennard-Jones pair potential as the basis of the training set. The training set is a set of boxes randomly populated with atoms. The potential energy of each box is calculated by summing the the Lennard-Jones pair potential function over all pairs of atoms within a specified cut-off distance. The list of pairwise distances for each box together with the associated box energies serve as a basis for the fitness function, that is in turn used to evaluate the fitness of the trees in the algorithm. The details of the training set and fitness function construction can be found in section \ref{subsec: TrainFit}, while the overall algorithm is described in section \ref{sec: Method}.

The Lennard-Jones functional form, which is the target function, minimally appears as a four-level tree in our representation {this is wrong}. The space of all possible trees of this size is very large (see section \ref{sec: Discussion}). The energy landscape is rough, discontinuous, and littered with local minima. We have developed a powerful optimization method which is a hybrid between a GA and Parallel Tempering Metropolis to enable efficient search on a massively parallel cluster architecture. The method evolves a set of populations of trees according to a scheme that combines mutation with a Metropolis Monte Carlo algorithm.  Each population is assigned a different effective temperature and trees are periodically exchanged between populations. The details of the algorithm can be found in section \ref{subsec: PT}.

Searching for several hours on $100$ processors, the algorithm reproducibly discovered the Lennard-Jones functional form, after traversing only a very small fraction of the total search space. We believe that this demonstrates the effectiveness of this algorithm as a method for unsupervised development of force-field functional forms. We believe that this is a first demonstration of unsupervised force-field functional form fitting. For more details, we refer the reader to section \ref{sec: Discussion}.

\section{Method} \label{sec: Method}

\begin{figure}[h] 
   \centering
   \includegraphics[height=1.5in]{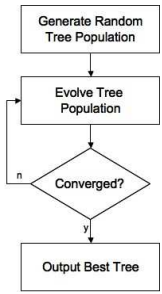} 
   \caption{\it \small Top Level Method Flowchart. Iterative refinement of a population of composite functions.}
   \label{fig:method}
\end{figure}

In general terms, our method iteratively refines an overall population of candidate GP trees with a convergence criterion defined by the fitness of the best individual tree in the population (Fig. \ref{fig:method}). The overall population is randomly generated at the start and is iteratively refined using our Parallel Tempering Genetic Program method (section \ref{subsec: PT}). 

The trees are permuted using a set of tree evolution operators, as described in section \ref{subsec: GP}, and move to the next generation according to their fitness, described in section \ref{subsec: TrainFit}. The "fittest" individuals are monitored to detect convergence. Once convergence is obtained, the best individual tree is output.

\subsection {Functional and Genetic Programming}   \label{subsec: GP}
Functional Programming \cite{hughes:matters-cj, AllisonL:Modmla} is a programming style that approaches computation as a hierarchical evaluation of functions. The concept of a {\it first-order function}, meaning a function that operates on functions, was defined in this context, though it has been widely used before without formal definition. For example, a {\it derivative} function takes one function to another function. In this formal setting, Genetic Programming is a case of a {\it first-order function}, that typically operates on {\it pure} (no side-effects) functions to construct and refine composite computer programs.

\begin{figure}[h]
   \centering
   \includegraphics[height=1.5in]{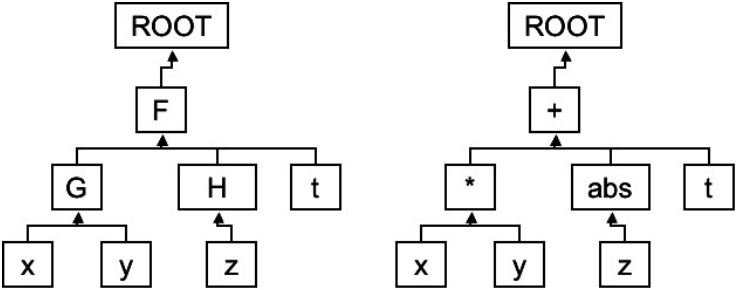} 
   \caption{Left: Graphical tree representation of a Functional Program F(G(x,y),H(z),t). The arrows indicate the data flow sense. Right: A particular instance of the program xy+abs(z)+t.}
   \label{fig:AbstractTree}
\end{figure}

Genetic Programming methodology was originally developed by Koza \cite{KozaJR:Genprh} to enable automatic generation of computer programs. The high-level strategy builds a population of random programs, computationally represented as nested {\bf trees} of pure functions, as depicted in Figure \ref{fig:AbstractTree}, and then iteratively refines this population using a GA and a fitness function that operates on a single tree. The refinement relies on existence of mutation {\it first-order} operators that change the structure of a tree. Such operators work on a tree either locally (mutation), or more drastically (crossover). Though the terminology for the mutation operators is borrowed from Genetic Programming, where such operators are reasonably well-defined, their description is much fuzzier in the case of a function tree.

Since we are attempting to build an algebraic expression for a force-field, all our pure {\bf tree elements} are unary or binary algebraic functions. A composite tree, traversed in the data-passing sense, from the leaves to the root, simply evaluates an algebraic expression. However, in general, as well as in our code, the pure operators are not so constrained. The library of such operators can consist of logical, calculus, transform, and any composite operators that can be encapsulated into a functional form.

The algebraic trees, that represent the functional form of the force-field become the object of subsequent optimization. The optimization is based on an importance sampling Metropolis Monte Carlo \cite{MetropolisN.:Equscb}. This is an iterative process that attempts to improve a population of trees by evolving a new "better" population in each refinement step. Evolution happens in three stages: generation, mutation, and testing.  The first stage produces $N$ new trees from $N$ old trees.  The second stage randomly mutates some of these trees. In the testing stage, we compare the fitness of each of the $N$ new trees with one of the old trees and pick one or the other based on the Metropolis acceptance probability.

In the generation stage, each new tree is created by either by pass-through or crossover, with equal probablility.  Pass-through involves selecting the fittest tree that has not already been passed though from the old population and copying it into the new population. Cross-over involves creating a new tree by combining two parent trees selcted from the old population.  Tournament selection is used to choose parent trees from the old population: four trees are selected with equal probability from the old population, and the tree with highest fitness is selected.  To perform the actual crossover operation, a depth level from the first parent is selected, with the restriction that the node is not the root (which would not mix up the trees at all) or at maximum depth (which would not cause enough change).  A depth level is then selected from the second parent, with the same restrictions.  A randomly chosen subtree rooted at the selected depth level is then chopped from the first parent, and replaced with a randomly chosen subtree rooted at the selected depth level in the second parent, producing one new child tree containing parts of both parents.  An additional restriction is that the child tree must satisfy the minimum and maximum allowable tree depths.
 
In the mutation stage, each new tree is either mutated or left alone with equal probability, without regard for fitness or how the tree was created.  On a tree that is mutated, a node is selected, which can include the root or a leaf.  The node is selected with equal probability from all nodes, meaning that there is a higher probability to select a node near the leaves than to select a node near the root; this was done to give a preference for small adjustments to the parameters rather than drastic changes to the entire functional form.  The subtree rooted at this node is chopped, and a new random subtree is made from scratch, restricted to keeping the entire tree within legal depth limits.  Thus, on average 1/4 of the new trees are made by crossover, 1/4 by crossover and mutation, 1/4 by mutation alone (pass-through followed by mutation), and 1/4 are retained from the previous generation.
 
In the testing stage, the old trees (which are ordered by fitness) and the new trees (which are in the order they were created, largely random with regards to fitness) are compared  head-to-head.  Metropolis compares each old tree against the new tree in the same position in the list. This is a simple variant of Tournament Selection. The new tree is accepted into the population if it is better then an old one, and only occasionally otherwise according to the Boltzmann probability:

\begin{eqnarray}
P_{acc}=min\{1, exp[(\beta(F_{new}-F_{old})]\} \\
\beta=\frac{1}{T}
\end{eqnarray} 

\subsection {Parallel Tempering} \label{subsec: PT}

Parallel Tempering (PT) was originally introduced by Swendsen and Wang \cite{SWENDSENRH:RepMCs} to deal with the local traps of spin glass energy surface. Their technique uses $N$ replicas of the system each at a different temperature and exchanges partial state information between replicas. The fundamental idea is to use the high-temperature replicas to sample the system phase space at a coarse level with the low-temperature replicas refining the states in local traps. In this way, a hybrid of local and global sampling is achieved. Later changes to the method replace partial information exchange with a complete state swap. Many parameters of the method have come under scrutiny since. For a review of the recent developments, see \cite{EarlDJ:Partta}.

\begin{figure}[h]
   \centering
   \includegraphics[height=.3in]{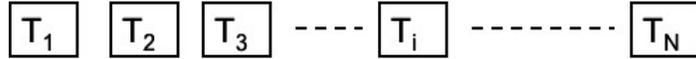} 
   \caption{Graphical representation of a Parallel Tempering algorithm with replicas marked with their individual temperatures.}
   \label{fig:Temper}
\end{figure}

 In our method, a single replica is a population of candidate trees. The replicas exchange information with their nearest neighbors in the temperature space (Figure \ref{fig:Temper})  by swapping random (or selected) trees with a probability based on the tree's relative Boltzmann weights with $F_i$ as the fitness of the tree selected in replica $i$ at a temperature $T_i$:

\begin{eqnarray}
P=min\{1, exp[(\beta_1-\beta_2)(F_1-F_2)]\} \\
\beta_i=\frac{1}{T_i}
\end{eqnarray}

We initialize the initial temperature distribution for replicas as either linear or logarithmic between upper and lower bounds. We can also allow for a dynamic temperature adjustment using acceptance/rejection ratios as a guide to perturbing individual temperatures.

\subsection {Training Set and Fitness Function} \label{subsec: TrainFit}
\subsubsection{Training Set} \label{subsubsec: Training}
The training set is a series of $N_{box}$ 3-dimensional domains (boxes) randomly populated with atom positions. Each box has an associated box energy which depends on the positions of all the atoms in the box.  In the current work, the box energy is given by a sum of pair potential functions, evaluated for all pairs of atoms lying within a given cut-off distance.  Periodic boundary conditions are used and pairs of atoms closer than a minimum interaction distance are ignored.  The pair potential was chosen to be the well-known Lennard-Jones potential function \cite{Lennard-JonesJE:Coh}.

\begin{eqnarray} \label{eqn:LJ}
E_{box}=\sum _i \sum_ {j>i} e_{i,j} \label{eqn:LJ-ej} \\
e _{i,j} = 4 \epsilon \{ {(\frac{\sigma}{R_{i,j}})}^{12} - {(\frac{\sigma}{R_{i,j}})}^6 \} \\
R_{i,j} =  \| {\vec{R_i} - \vec{R_j}} \| 
\end{eqnarray}

where $\epsilon$ and $\sigma$ are energy and length parameters.

The procedure produces $N_{box}$ box energies, one for each box. The set of the atomic coordinates for each box with periodic boundary conditions, and the corresponding set of $k$ surrogate target box energies become the training set.

\subsubsection{Fitness Function} \label{subsubsec: Fitness}

Fitness of a tree is directly based on the training set of $k$ boxes. The fitness function for a tree is evaluated by computing the total energy for each of the boxes using that tree,

\begin{eqnarray} \label{eqn:TreeE}
\tilde{E} _{box}=\sum_i \sum _{j>i} \tilde{e}_{i,j} \label{eqn:LJ-etilde} 
\end{eqnarray}

where $\tilde{e} _{i,j}$ is the pair potential function represented by the tree. The fitness of the tree is then given by

\begin{eqnarray} \label{eqn:LeastSq}
F_{tree}^2 = (-\frac{1}{k_{box}})\sum _{box=1} ^k{(E_{box} - \tilde{E}_{box})}^2,
\end{eqnarray}

where the minus sign is required to have increasing fitness correspond to decreasing error.

\section {Results} \label{sec: Results}
The purpose of this study was to test whether the GP approach was capable of discovering accurate potential energy functions that replicate the true potential energy surface. Typically, information about the true potential energy surface is obtained by using quantum mechanics calculations to evaluate the energy of small configurations of atoms.  Hence, we chose as our test case 10 boxes containing 10 randomly positioned particles. In order to construct a problem with a known global optimum, the energies of the boxes were calculated using the Lennard-Jones pair potential, as described above.  Each box had dimensions of $3 \sigma \times 3 \sigma \times 3 \sigma $. The particles were placed randomly in the box, but no particles were allowed to come closer than $ 0.5 \sigma $  All pairs distances in the range $ 0.7 \sigma < R < 2.0 \sigma $ were recorded and used to compute the target box energy, taking periodic images into account. The values of $ \sigma and \epsilon$ were set to unity. This resulted in about 60 pair distances per box. By varying the random number seed, we generated four independent test cases.  For each test case we executed two different parallel tempering optimizations, with either $N = 10,000$ or $N = 50,000$ individual trees in each population.  In all cases, we used 200 populations with temperatures distributed logarithmically from 0.1 to 10 $\epsilon^2$ (the units of temperature are the same as those of the fitness function i.e. $\epsilon^2$).  All trees were required to have minimum depths of 3 and maximum depths of 4. The calculations were run on a cluster of 100 AMD Opteron 2.2 GHz processors with Quadrics interconnects.  For the runs with 10,000 individuals per population, each generation required about 100 seconds. For 50,000 indiviuals, the time per generation was about 5 times longer. Most of this time was spent in the evaluation of box energies.

\begin{figure}[htp]
  \centering
  \subfigure[Lennard-Jones equivalent tree 1]{\label{fig:edge-a}\includegraphics[scale=0.3]{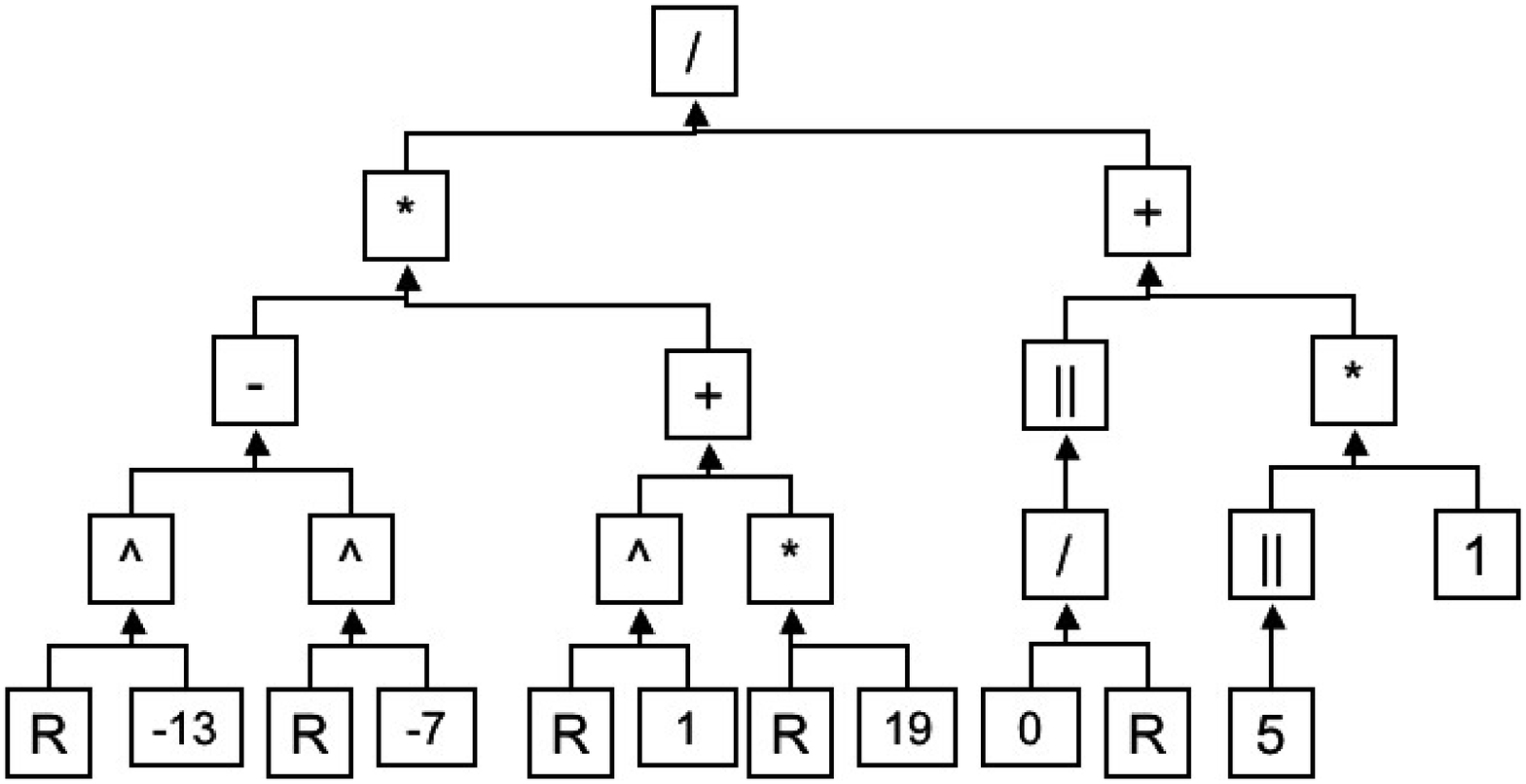}} \ \ \  \  \
  \subfigure[Lennard-Jones equivalent tree 2]{\label{fig:edge-b}\includegraphics[scale=0.3]{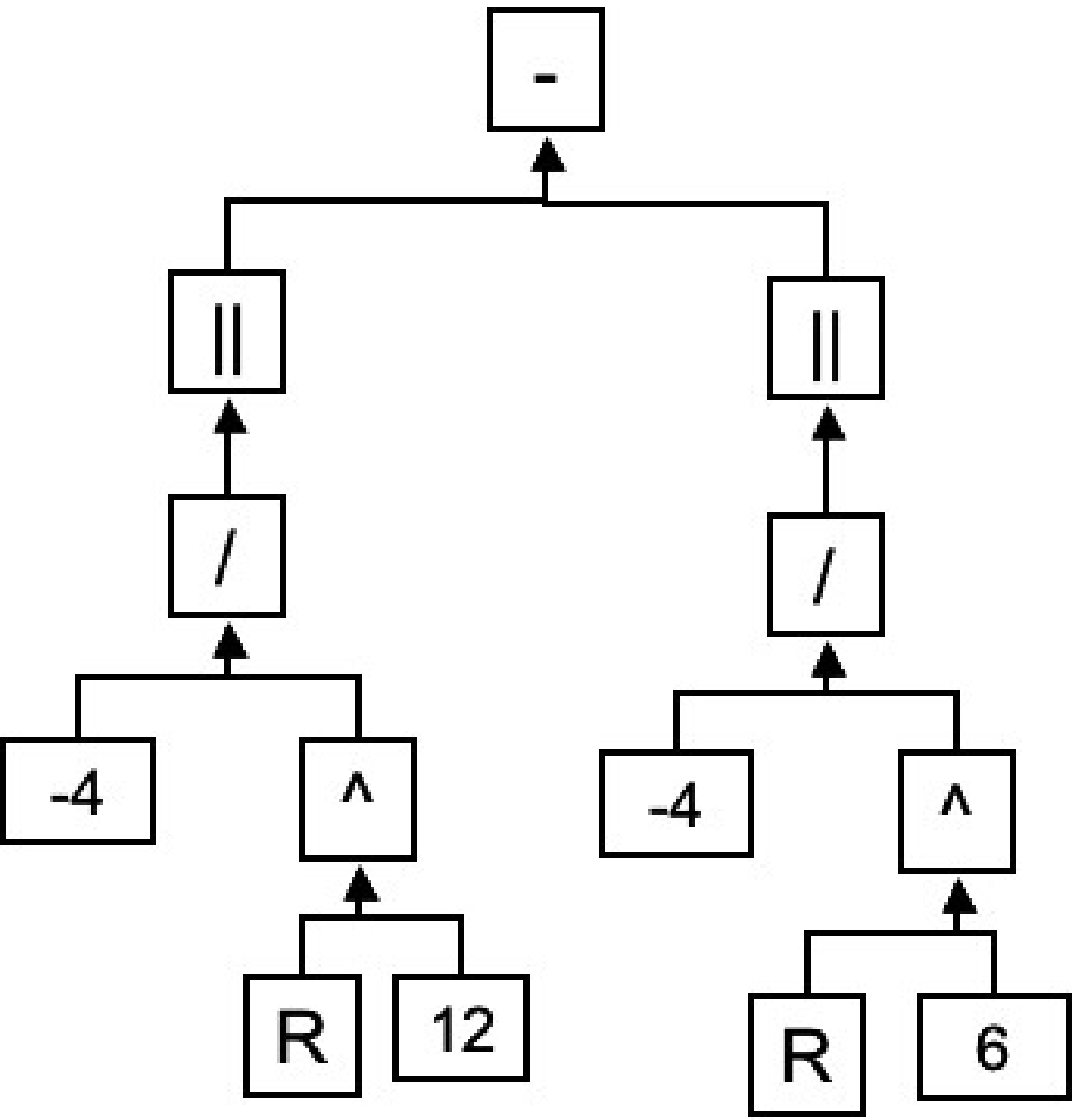}} \\
  \subfigure[Lennard-Jones equivalent tree 3]{\label{fig:edge-c}\includegraphics[scale=0.3]{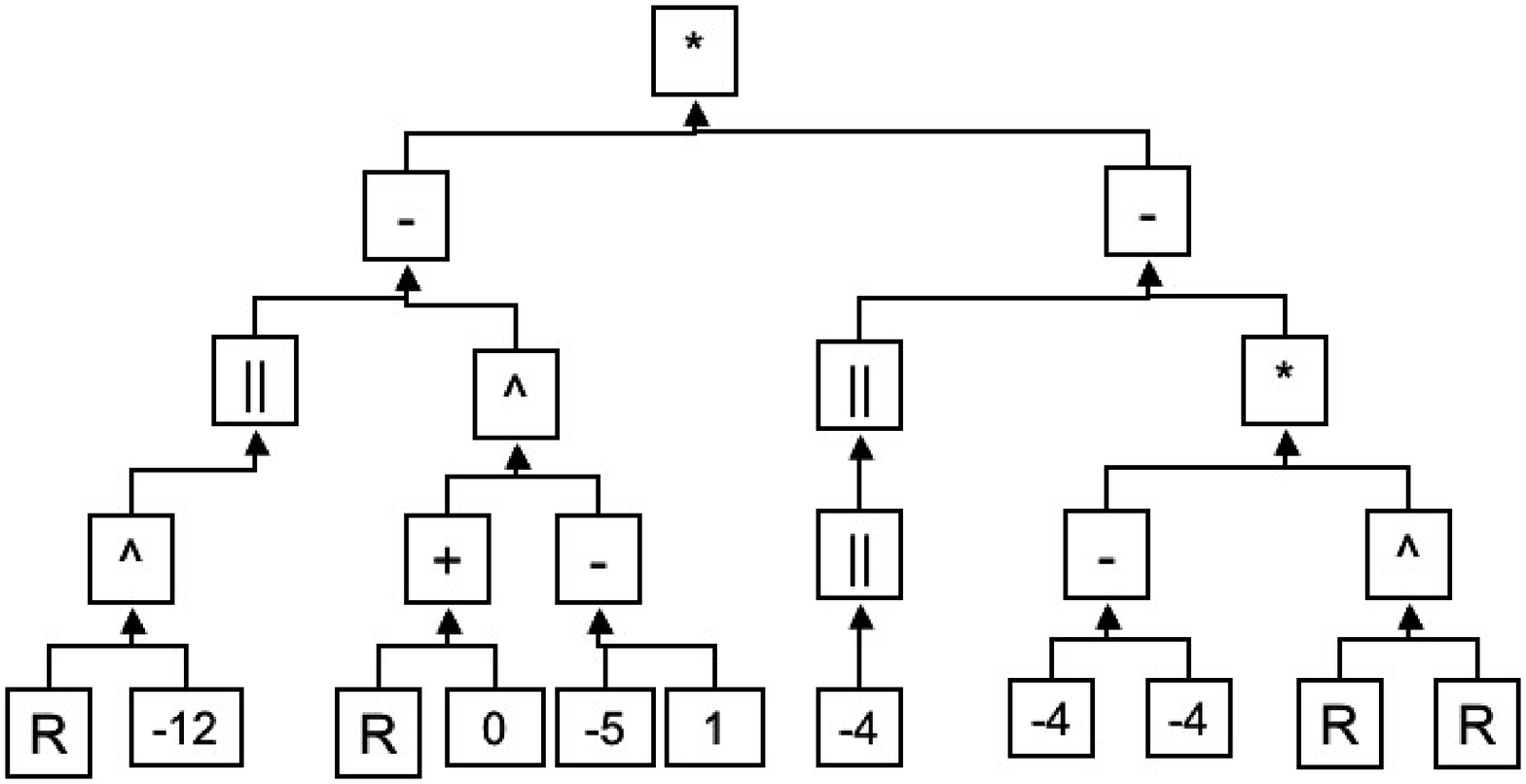}} \
  \subfigure[Near-miss tree with high fitness value.]{\label{fig:edge-d}\includegraphics[scale=0.3]{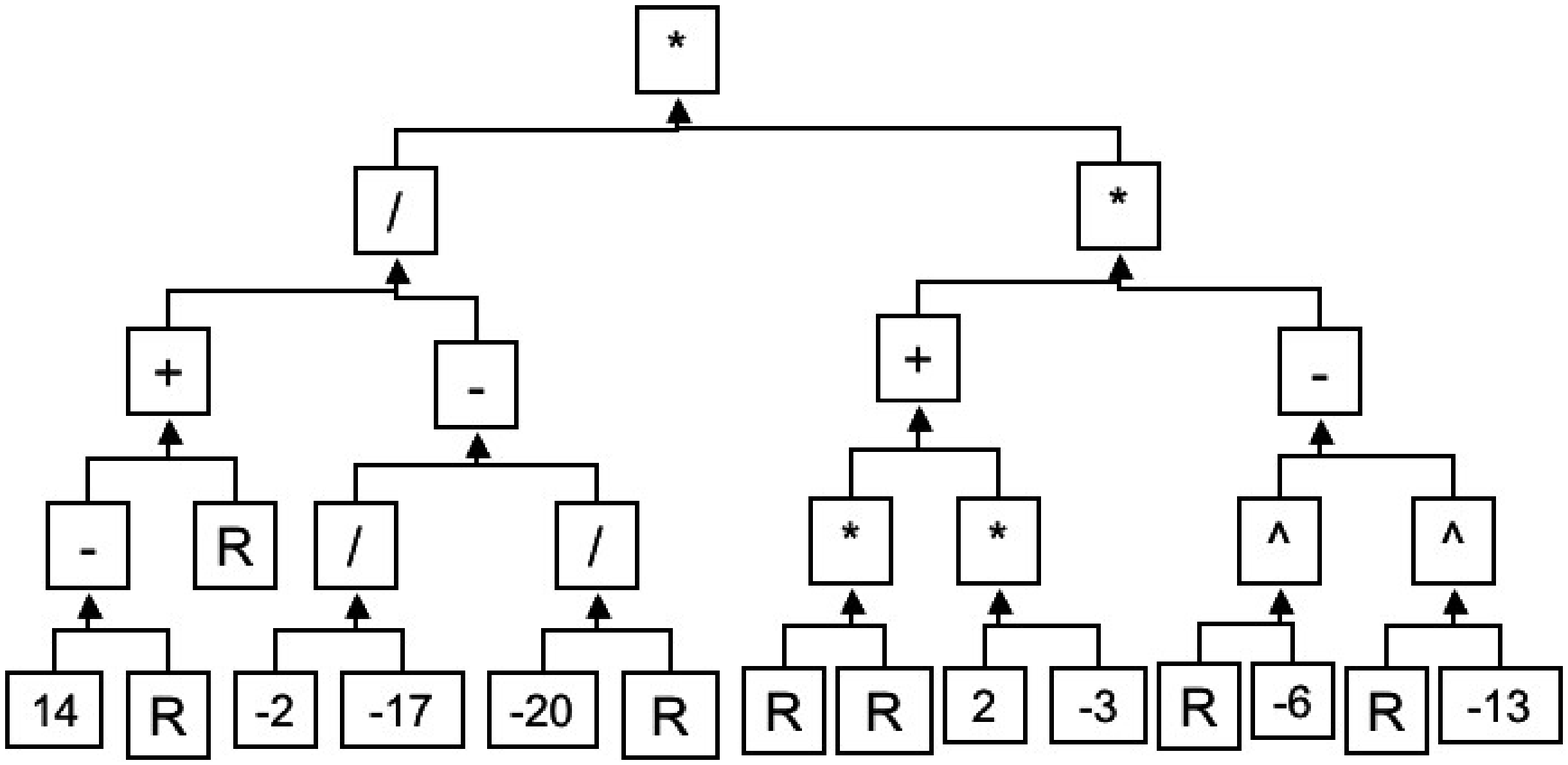}} \
 \caption {Lennard-Jones equivalent trees and a near miss. The trees shown in (a), (b), and (c) are algebraically equivalent to the Lennard-Jones form. The tree shown in (d) produces a function that replicates Lennard-Jones numerically over the range of interest to with error less then 10\%. For example, the function represented in (a) is $\frac{[R^{-13}-R^{-7}][R^1+19R]}{\| \frac{0}{R} \| + \|5 \| \times 1}$, (b) literally is $\| \frac{-4}{R^{12}} \| - \| \frac{-4}{R^{6}} \| $, and (c) is given by $ (\|R^{-12}\| - (R+0)^{-5-1}) \times (\|\|-4\|\| - (-4-(-4) \times R^R) $, all of which reduce to $4(R^{-12}-R^{16})$. The tree (d) is a "good" approximation to the above function.}
  \label{fig:LJTrees}
\end{figure}

The results of the runs are stochastic, both in the initial conditions and the optimization method, and so we see a distribution of behaviors. However, most runs successfully found an algebraic equivalent of the original target function.  Algebraic equivalence here means that the functional form can be transformed to the exact Lennard-Jones form by a sequence of algebraic transformations. Three such algebraic equivalents are displayed in Figure \ref{fig:LJTrees}. 

\begin{figure}[htp]
  \centering
  \subfigure[Average square error for a series of runs.]{\label{fig:runs-a}\includegraphics[scale=0.37]{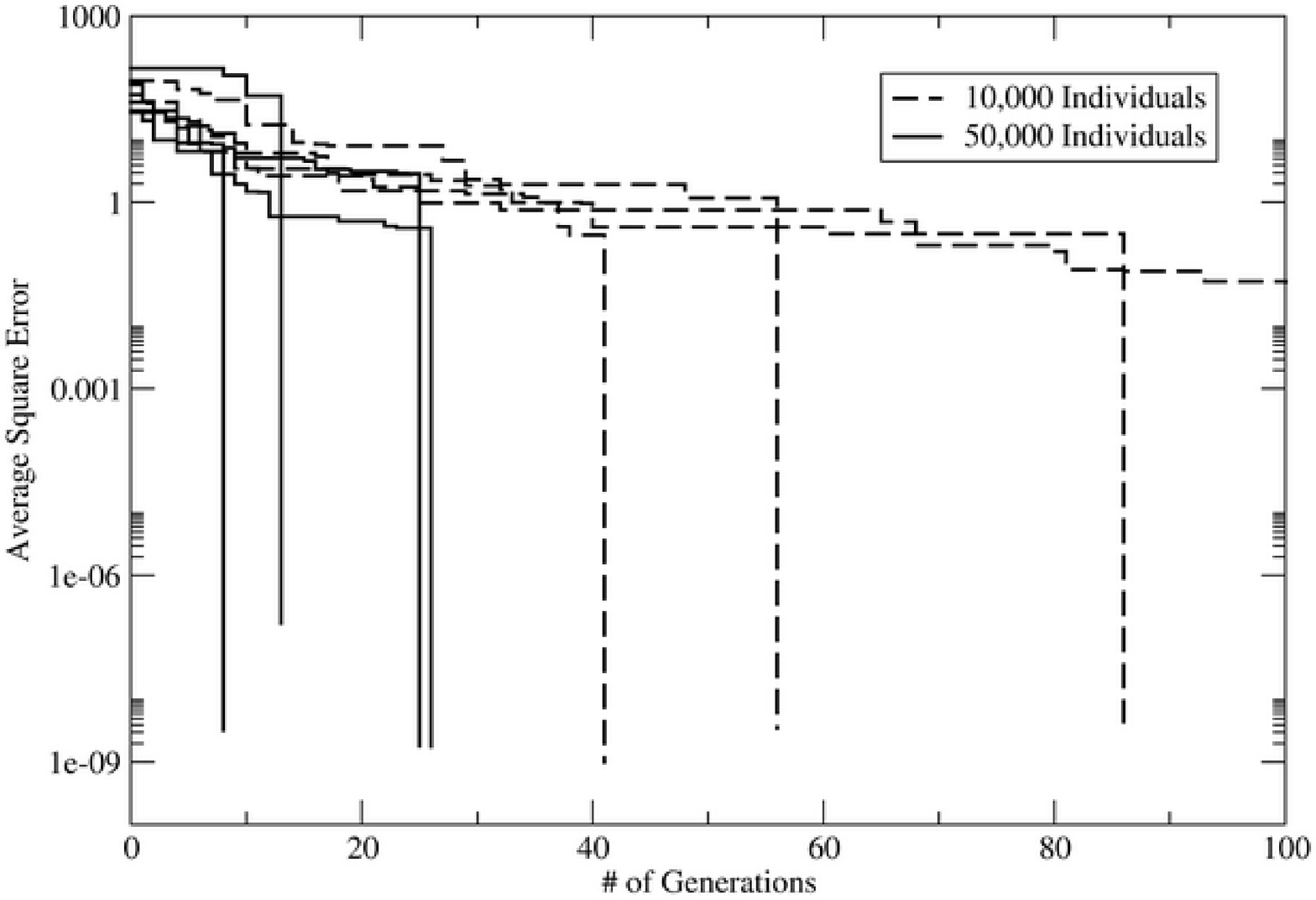}} \ \ \
  \subfigure[Average square error for a single run.]{\label{fig:runs-b}\includegraphics[scale=0.37]{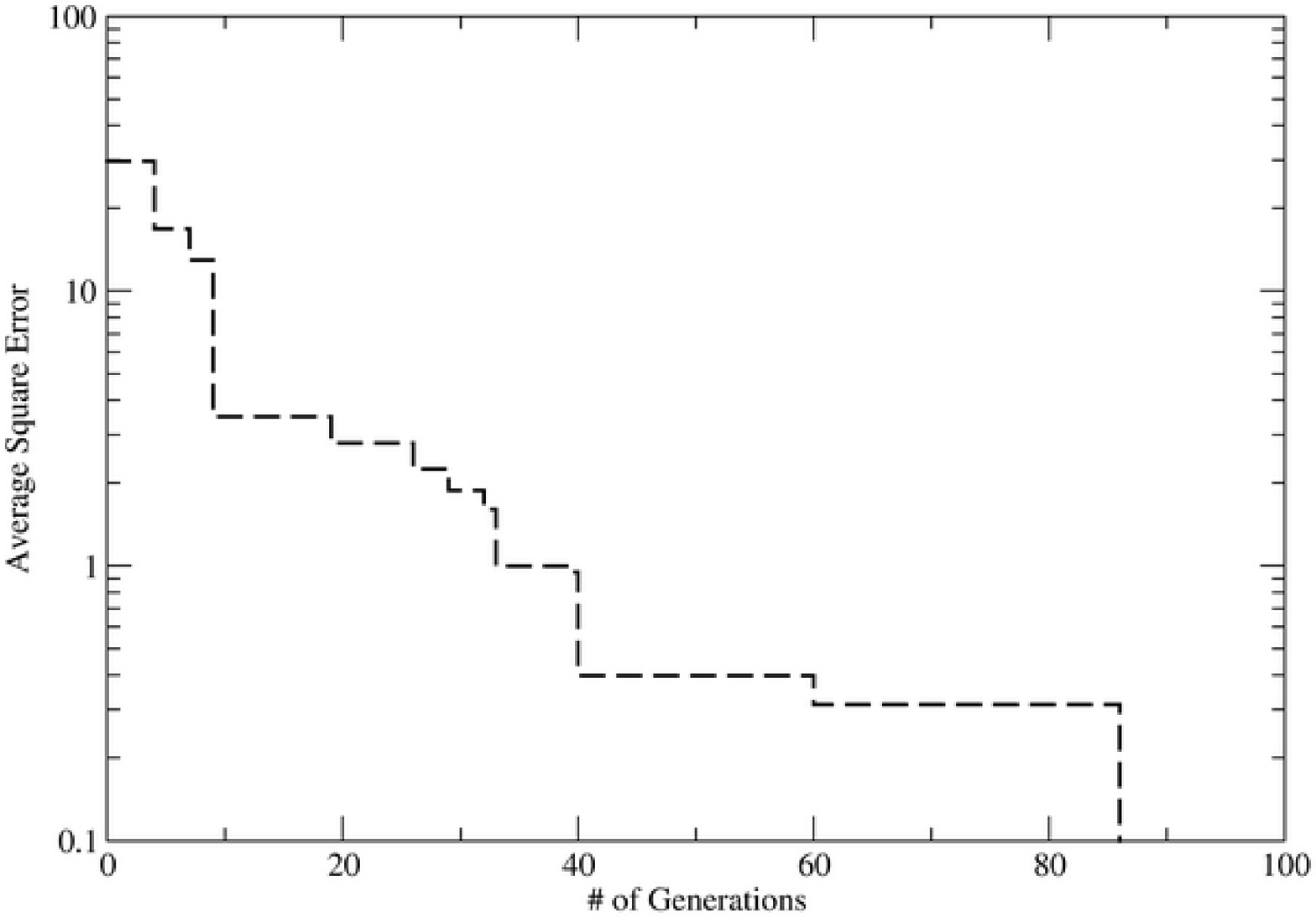}} 
  \caption {Convergence of the average square error (negative fitness) of the overall fittest tree after each generation, (a) all eight runs, which in most cases successfully found algebraic equivalents of the Lennard-Jones function, (b) One of the runs in more detail.}
  \label{fig:Runs}
\end{figure}

  Fig. \ref{fig:runs-a} shows the average square error for the overall fittest tree for
each generation. A total of 8 runs are shown. The dashed lines indicate four
independent runs with $N = 10,000$ individuals in each population, each
using different initial populations and different test boxes.  The solid
lines indicate four independent runs with $N = 50,000$ individuals in each
population, each using different initial populations and the same  test
boxes used for the first four runs.

Of the four independent runs with $N = 10,000$, three of them successfully
found algebraic equivalents of Lennard-Jones function. The residual average square error of
approximately $10^{-9} \epsilon^2 $ can be attributed to machine error. The
fourth run failed to find an algebraic equivalent, even after 400 generations.
However, it did find several functions which are good approximations to the 
Lennard-Jones function, but have quite different functional forms e.g. the tree shown in Figs. \ref{fig:edge-a}, \ref{fig:edge-b}, and \ref{fig:edge-c} .

In the case of the larger populations, $N = 50,000$, all four runs found algebraic equivalents, and 
did so in substantially fewer generations. Clearly, the greater diversity of these larger populations improves the
search efficiency.

Fig. \ref{fig:runs-b} shows more clearly how one of the $N= 10,000$ runs progressed.  Initially fitness improves quite steadily, 
until a good approximation to the exact Lennard-Jones function is found.  After this point, further improvement occurs only sporadically.
Eventually, the algebraic equivalent appears.

\section {Discussion} \label{sec: Discussion}

The problem of finding the correct tree in the space of all possible trees of a given size is made difficult by the sheer size of the search space. Our algorithm, starting with an initial total populations of $\sim 10^7$ trees, in most cases found the global optimum in less than $100$ generations. This provides an upper bound for a number of trees surveyed of $10^9$. Compare this figure to the number of possible trees of depth $4$. 

We will ignore the unary operator $\| \|$, and ignore are trees that are not maximal. Then, for $M$ {\it binary} operators, and a maximum depth of $K$, the number of possible trees at operator-only level is given by:
$$
N_{op}=M \times M^2 \times ... M^{2^{K-1}}=\prod_{k=0}^{K-1} {M^{2^k}}.
$$
The number of leaves holding integer constants on the range $[-P,P]$ or the input variable is given by 
$$N_{val}=(2P+2)^{2^K}.$$ 

In our case $K = 4, M = 5, and P = 20$, and so the total number of possible trees is, 
$$
N_{tree} = N_{leaves} \times N_{op} = 42^{16} \times 5^{15} = 2.8 \times 10^{36} 
$$

Hence we find that an upper bound on the ratio of the total space to the searched space is roughly $10^{27}$.  Clearly, this is a non-trivial problem.  For this reason, it is not surprising that we appear to be the first group to attempt an unsupervised automated approach to finding functional forms for force-fields.  Previous efforts at unsupervised force-field fitting have been restricted to parameter optimization, given a fixed functional form. We succeeded because we were able to use relatively large computational resources, and we used a very robust optimization method.  

It is important to emphasize that while the surrogate test problem used in this study had a known solution, it nonetheless was representative of many potential energy surfaces where accurate functional forms are unknown.  We deliberately chose a training set that closely resembled data generated by quantum density functional theory calculations of potential energy.  Given the success of the method in the current study, we now intend to apply the method to some of the many atomic systems for which existing force-fields have been found lacking.

\bibliography{force_field}
\end{document}